\documentclass{article}
\usepackage{spconf,amsmath,graphicx}
\usepackage{multirow}
\usepackage[colorlinks, linkcolor=blue, citecolor=blue]{hyperref}

\title{A DEEP GRADIENT BOOSTING NETWORK FOR OPTIC DISC AND CUP SEGMENTATION}
\name{Qing Liu$^{\star}$ \qquad Beiji Zou$^{\star}$ \qquad Yang Zhao$^{\star}$ \qquad Yixiong Liang$^{\star}$}
\address{$^{\star}$School of Computer Science, Central South University, Changsha 410083, P.R. China}
\begin{document}
%
\maketitle
\begin{abstract}
Segmentation of optic disc (OD) and optic cup (OC) is critical in automated fundus image analysis system. Existing state-of-the-arts focus on designing deep neural networks with one or multiple dense prediction branches. Such kind of designs ignore connections among prediction branches and their learning capacity is limited. To build connections among prediction branches, this paper introduces gradient boosting framework to deep classification model and proposes a gradient boosting network called BoostNet. Specifically, deformable side-output unit and aggregation unit with deep supervisions are proposed to learn base functions and expansion coefficients in gradient boosting framework. By stacking aggregation units in a deep-to-shallow manner, models\rq ~performances are gradually boosted along deep to shallow stages. BoostNet achieves superior results to existing deep OD and OC segmentation networks on the public dataset ORIGA.
\end{abstract}
\begin{keywords}
Fundus image, OD and OC segmentation, gradient boosting, deep supervision
\end{keywords}
\section{Introduction}
\label{sec:intro}
\begin{figure}[!b]
	\centering
	\includegraphics[width=0.7\linewidth]{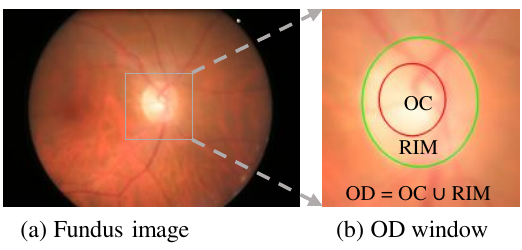}
	\caption{Structure illustration for optic disc (OD) in retinal fundus image. It includes two parts: optic cup (OC) and neuralretinal rim. (a) A fundus image from ORIGA \cite{ORIGA}. (b) OD window cropped from (a). }
	\label{fig:exmaple_OD_OC}
\end{figure}
Glaucoma is the leading cause of irreversible blindness. It is a chronic eye disease that degrades the optic nerves, resulting in a large ratio between optic cup (OC, see Fig. \ref{fig:exmaple_OD_OC}) to optic disc (OD, see Fig. \ref{fig:exmaple_OD_OC}) \cite{CDR_IOVS_2000}. In clinical, the cup-to-disc ratio (CDR) is manually estimated by expertised ophthalmologists. It is labour intensive and time-consuming. To make the accurate CDR quantification automated and assist glaucoma diagnosis, the segmentation of OD and OC is attracting a lot of attention.

\begin{figure}[!t]
	\centering
	\includegraphics[width=\linewidth]{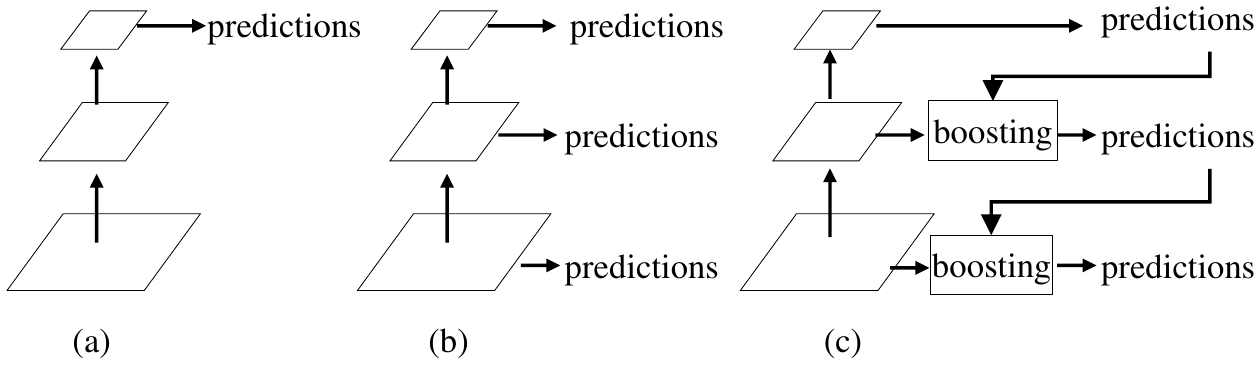}
	\caption{Architecture comparison. From left to right: (a) Single dense prediction branch. (b) Multiple dense prediction branches. (c) Multiple dense prediction branches with boosting modules. The boosting modules are stacked in a deep-to-shallow manner and learn residuals to boost previous classification model.}
	\label{fig:architecture_compare}
\end{figure}
\begin{figure*}[!t]
	\centering
	\includegraphics[width=0.8\linewidth]{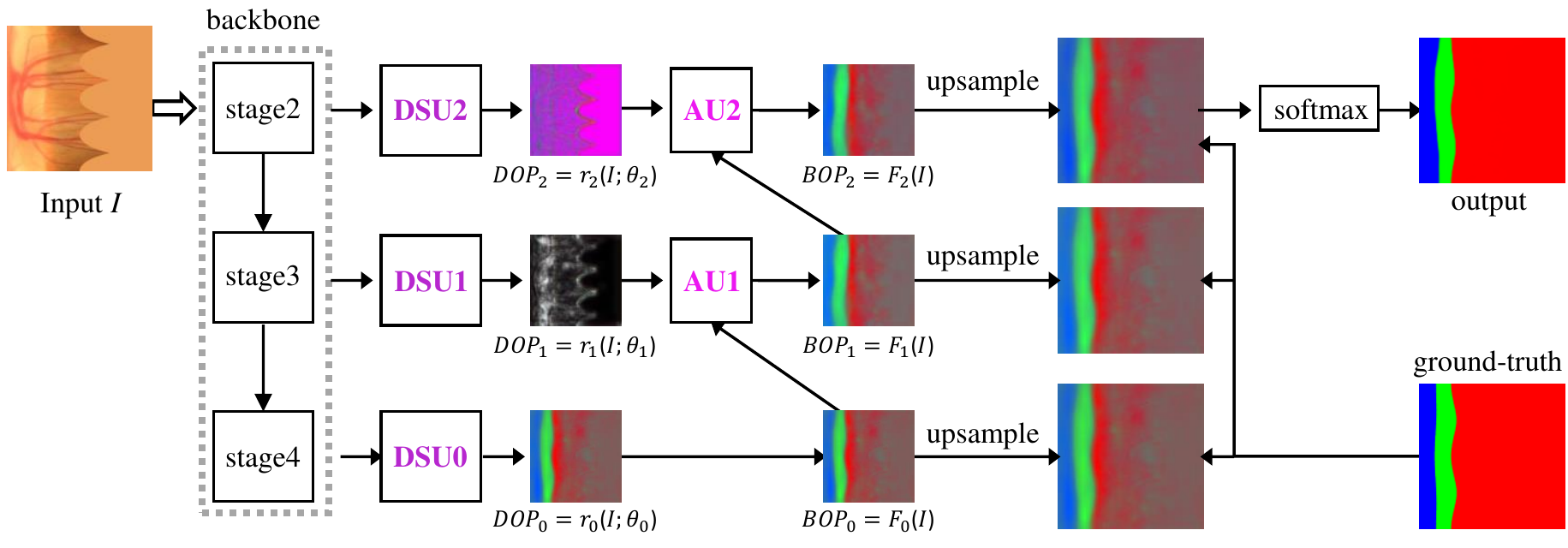}
	\caption{The architecture of proposed BoostNet. The deformable side-output units (DSU) and aggregation units (AU) with deep supervision learn residuals and the expansion coefficients to boost the dense classification models. The implementation details about DSU and AU are illustrated in Fig. \ref{fig:DSU_AU}. DSU together with AU build the boosting module in Fig. \ref{fig:architecture_compare} (c). Input is a polar OD window, which is same with previous methods MNet \cite{Fu_TMI_2018} and SAN \cite{Liu_NC_2019}. }
	\label{fig:DHRNet}
\end{figure*}
\begin{figure*}[!h]
	\centering
	\includegraphics[width=0.8\linewidth]{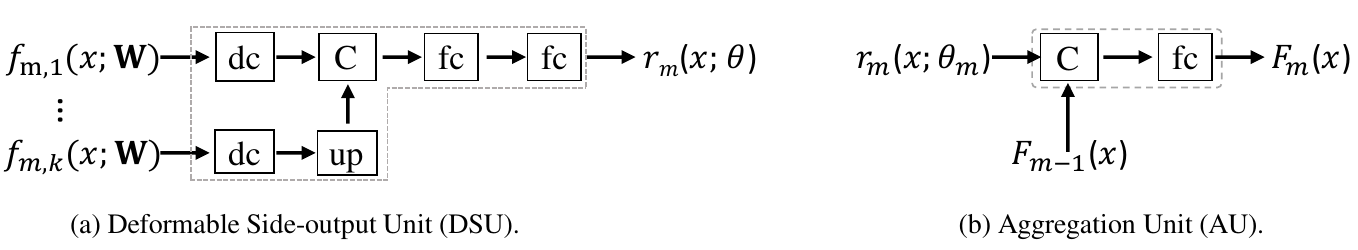}
	\caption{Implementation for (a) DSU and (b) AU. \lq dc' and \lq fc' denote deformable convolutional layer \cite{deformableCNN} and full connection layer respectively. \lq C\rq~ and \lq up' denotes concatenation and upsample operators respectively.}
	\label{fig:DSU_AU}
\end{figure*}

\indent Current state-of-the-arts for OD and OC segmentation are data-driven. It is formulated as a dense classification problem and solved by designed deep neural networks consisting of a feature learning network and one or multiple prediction branches. Some works such as SAN \cite{Liannan_NC_2018} and CE-Net \cite{gu2019cenet} directly append one branch of prediction on the final output of a CNN architecture, as is shown in Fig. \ref{fig:architecture_compare} (a). This facilitates for recognition since high-level features contain rich semantic information. However, they ignore the spatial details in shallow layers, which results in inaccurate classification around edges. To make use of spatial details, others such as MNet \cite{Fu_TMI_2018} and AG-Net \cite{zhang2019AGnet} make predictions on multiple levels of features. The final outputs are obtained by element-wise summation. Fig. \ref{fig:architecture_compare}(b) illustrates the architecture with multiple prediction branches. However, in this kind of design, multiple prediction branches are independent, which limits the learning capacity of the segmentation model. Regarding each prediction branch as a classification model, how to ensemble them for better OD and OC segmentation remains unsolved. In the field of traditional machine learning, an immediate thought for model ensemble is boosting. This motivates us to solve the dense classification problem in a gradient boosting framework \cite{boosting} \cite{saberian2011multiclass} and develop an architecture with boosting modules to boost the branch's classification ability. Fig. \ref{fig:architecture_compare}(c) illustrates the proposed boosting idea.


\indent In detail, this paper explores a boosting deep neural network called BoostNet for OD and OC segmentation. It learns a sequence of base functions with deformable side-output unit (DSU) to boost the dense classification model in a stage-wise manner. Each base function learns the residual between ground-truth and output by classification model at the previous stage. Combining the classification model at the previous stage with base function in an additive form by the proposed aggregation unit (AU) results in a better classification model. Stacking the AUs with deep supervision in a deep-to-shallow manner, the classification models are boosted stage-by-stage. Fig. \ref{fig:DHRNet} illustrates the architecture of our BoostNet. We note that our BoostNet is different from existing combination of boosting and CNNS called BoostCNN \cite{moghimi2016boosted}. Our BoostNet is designed for pixel-level classification which treats the inner sub-networks as weaker learners while BoostCNN \cite{moghimi2016boosted} is proposed for image-level classification which treats the classification models after different training iterations or trained with different inputs as weaker learners.

\indent The contributions of this paper are two-fold: (1) We introduce gradient boosting into convolutional neural network and design BoostNet for OD and OC segmentation; (2) We show that BoostNet achieves state-of-the-art performances on public OD and OC segmentation dataset ORIGA \cite{ORIGA}.


\section{Gradient BOOSTING NETWORK}
\label{sec:dsanet}
The proposed gradient boosting network (BoostNet) roots in gradient boosting \cite{boosting} and is equipped with the designed Deformable Side-output Units (DSUs) and Aggregation Units (AUs). It can be trained in an end-to-end way. Similar to previous methods MNet \cite{Fu_TMI_2018} and SAN \cite{Liu_NC_2019}, BoostNet takes polar images as input and outputs polar segmentation maps. 

\textbf{Problem Formulation.} The goal of OD and OC segmentation is to find a function that maps the input ${x}$ to an expected label $y$ $\in\{oc, rim, bkg\}$ such that some specified loss function $\mathcal{L}$ over the joint distribution of all $(y, {x})$ is minimized:
\begin{equation}
\label{eq:eq1}
F^{\ast}({x}) = \arg \min_{F({x})} E_{{x}, y} \mathcal{L}(y, F({x})) \;.
\end{equation}
By merging the predicted OC and rim masks, OD is obtained. 

\indent Different from previous methods \cite{Liu_NC_2019} \cite{Fu_TMI_2018} \cite{zhang2019AGnet} \cite{ETNet} that directly estimate the function $F^{\ast}$ in Eq. \ref{eq:eq1}, we solve it from the view of gradient boosting \cite{boosting}.  The key thought is to approximate $F^{\ast}$ by greedily learning a sequence of base functions in an additive expansion of the form \cite{boosting}:
\begin{equation}
\label{eq:additiveExp}
F({x}) = \sum_{m=0}^{M} \beta_m r_m({x}; {\theta}_m) \;,
\end{equation}
where $r_0, \cdots, r_M$ are called base functions parametrised by ${\theta}_0, \cdots, {\theta}_M$ respectively and $\beta_0, \cdots, \beta_M$ are the expansion coefficients. Commonly, they are solved in a stage-wise way. First, the start base function $r_0({x})$ is optimised and $\beta_0=1$. The initial approximated function is determined by: 
\begin{equation}
\label{eq:startFunc}
F_0(x) = \arg\min_{r_0({x}; {\theta}_0)}E_{{x},y}\mathcal{L}(y, r_0({x}; {\theta}_0)) \;.
\end{equation}
Then for $m=1,\cdots,M$, the approximated function is expressed as:
\begin{equation}
\label{eq:baseFunc_m_add}
F_m({x}) =  F_{m-1}({x}) + \beta_m r_m({x}; {\theta}_m) \;.
\end{equation}
The base function $r_m({x}; {\theta}_m)$ and expansion coefficient $\beta_m$ are determined by:
\begin{equation}
\label{eq:baseFunc_m}
\arg \min_{r_{m}({x},{\theta}_m), \beta_m} E_{{x},y} \mathcal{L}(y, F_{m-1}({x}) + \beta_m r_m({x}; {\theta}_m))\;.
\end{equation}
We note that the base function $r_m({x}; {\theta}_m)$ here is the residual of $F_{m-1}(x)$. It is similar to the residual learning in ResNet \cite{ResNet} except that base functions in gradient boosting framework learn residuals of models\rq ~outputs while residual block in ResNet \cite{ResNet} learns the residual of features. By aggregating with residuals stage-by-stage under supervision, the models are boosted gradually and make predictions closer to ground-truth. Next we will detail how to design a network to learn the base functions and the coefficients.
 
\textbf{Gradient Boosting Network (BoostNet).} CNNs such as VGG \cite{VGG16}, ResNet \cite{ResNet}, DenseNet \cite{DenseNet} and HRNet \cite{HRNet} produce feature maps in a stage-wise manner. It is in line with the way that gradient boosting learns base functions and coefficients. This naturally motivates us to leverage the CNN to solve the optimisation problem defined by Eq. \ref{eq:startFunc} - Eq. \ref{eq:baseFunc_m}. 

Fig. \ref{fig:DHRNet} shows the proposed BoostNet. It consists of a backbone network, three deformable side-output units (i.e. DSU0, DSU1, DSU2) and two aggregation units (i.e. AU1 and AU2). The outputs of DSU0, DSU1, DSU2 are denoted as $DOP_0, DOP_1, DOP_2$ and outputs of AU1, AU2 are denoted as $BOP_1, BOP_2$ respectively. We adopt HRNet \cite{HRNet} as the backbone network since it is able to learn powerful representation with high-resolution. HRNet\cite{HRNet} consists of four stages. Each has multiple feature learning branches and produces multiple groups of features with different spatial resolutions. From first to fourth stage, the number of branches ranges from one to four. For brevity, only feature maps from second stage to the last are shown. The backbone network together with DSUs are used to learns base functions and AUs are used to learn coefficients. 

\indent Specifically, to start the boosting, a DSU called DSU0 is attached on the fourth stage of backbone network to produce $DOP_0$. Let $BOP_0 = DOP_0$ and with supervision on $DOP_0$, $F_0$ in Eq. \ref{eq:startFunc} is determined. Similarly, DSU1 is appended on third stage of backbone network to learn base function $r_1$. An AU called AU1 with supervision aggregates the $r_1$ and $F_0$ resulting in a boosted model $F_1$. By stacking AUs with supervision in a deep-to-shallow way, the models\rq~ performances are boosted gradually.

\textbf{Deformable Side-output Unit (DSU).} Fig. \ref{fig:DSU_AU}(a) shows the implementation of proposed DSU. It is attached on the end of a stage of the backbone network. Supposing the inputs of $m-$th DSU are $f_m(x)=\{f_{m,1}(x), \cdots,f_{m,k}(x)\}$ where $k$ is the number of feature learning branches, a deformable convolutional layer \cite{deformableCNN} with parameters $w^{dc}$ is first attached on each feature learning branch and the outputs are upsampled to same spatial resolution. Then those upsampled features are concatenated and followed by two full connection layer. The first $fc$ layer parametrised by $w^{fuse}$ fuses concatenated features. The second $fc$ layer parametrised by $w^{res}$ learns the residual. Denoting parameters in backbone network as $\mathbf{W}$, then parameters to be learned in a base function include $\theta = \{\mathbf{W}, w_{1}^{dc}, \cdots, w_{k}^{dc}, w^{fuse}, w_{res}\}$.

\textbf{Aggregation Unit with Deep Supervision.} The AU is designed to learn the expansion coefficient in Eq. \ref{eq:baseFunc_m_add}. Fig. \ref{fig:DSU_AU}(b) shows its implementation. It first concatenates the outputs by classification model at previous stage $F_{m-1}(x)$ and the residual by base function $r_m(x; \theta_m)$ at current stage, then a full connection layer with parameters is attached. With the supervision, coefficients are learned and a boosted segmentation model is obtained. In BoostNet, cross-entropy loss is used.

\section{Experimental Results}
To show the effectiveness of out proposed BoostNet, we validate it on the public dataset ORIGA \cite{ORIGA} and compare it with seven state-of-the-arts.

\textbf{Data Preprocess and Augmentation.} ORIGA dataset contains 650 pairs of fundus images and annotations by experts. Among ORIGA \cite{ORIGA}, 325 pairs are used for training and the rests for testing. Since we focus on OD regions, OD windows size of $640\times640$ are cropped from original images size of $3072\times2048$. Inspired by \cite{Polar_ICASSP_2018} \cite{Fu_TMI_2018} \cite{Liu_NC_2019}, a radial transform is first performed on OD windows in Cartesian coordinate system before feeding them into BoostNet. To increase the diversity of training data, three tricks are used to augment the training data in Cartesian system: (1) random OD windows cropping near the OD centre ( $\leq$ 20 pixels of horizontal and vertical offsets to OD centre); (2) multi-scale augmentation (factors ranging from 0.8 to 1.2); (3) horizontal flipping. For training data, the OD centres are determined based on the ground-truth OD masks. For testing data, we train an HRNet \cite{HRNet} for OD segmentation with the training data of ORIGA and estimate OD centres according to the predicted OD masks. During testing phase, vertical flipping augmentation is used on polar OD windows.

\textbf{Experimental Setting.} Our BoostNet is built on the top of implementation of HRNet \cite{HRNet} within the PyTorch framework. We initialize the weights in HRNet \cite{HRNet} with the pre-trained model on ImageNet \cite{imagenet} and the rest weights with Gaussian distribution with zero mean and standard deviation $0.001$. Parameters are optimised by SGD on three GPUs. Hyper-parameters includes: base learning rate (0.01, poly policy with power of 0.9), weight decay (0.0005), momentum (0.9), batch size (9) and iteration epoch (200).

\begin{figure}[!t]
	\centering
	\includegraphics[width=0.9\linewidth]{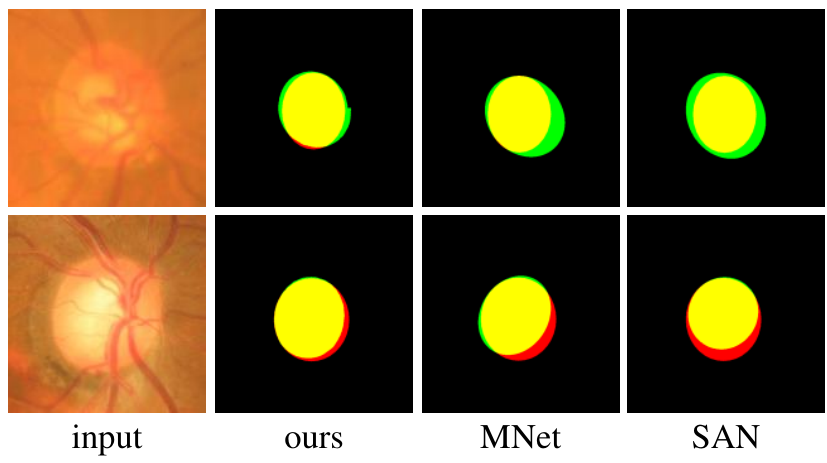}
	\caption{Examples for OC segmentation. From left to right are input OD window, results by ours, MNet \cite{Fu_TMI_2018} and SAN \cite{Liu_NC_2019}. Pixels in yellow, red and green are correct, miss and error detections respectively.}
	\label{fig:exmaple_OC_ORIGA}
\end{figure}

\begin{figure}[!t]
	\centering
	\includegraphics[width=0.9\linewidth]{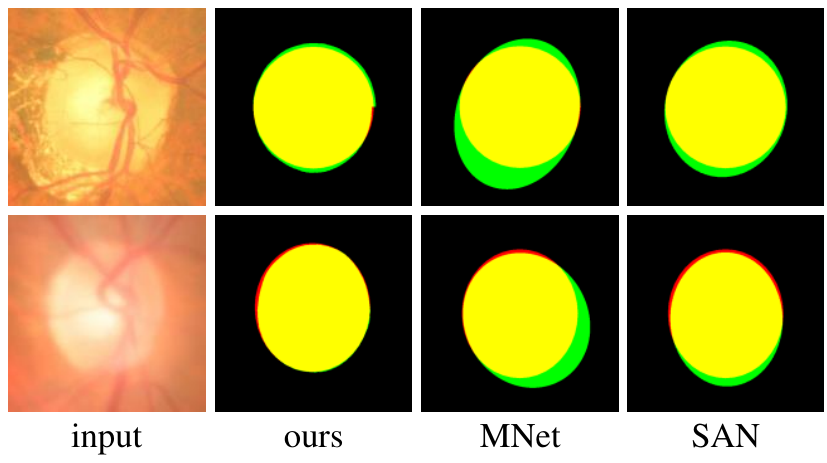}
	\caption{Examples for OD segmentation. From left to right are input OD window, results by ours, MNet \cite{Fu_TMI_2018} and SAN \cite{Liu_NC_2019}. Pixels in yellow, red and green are correct, miss and error detections respectively.}
	\label{fig:exmaple_OD_ORIGA}
\end{figure}

\textbf{Comparison with State-of-the-arts.} We compare the proposed BoostNet with seven state-of-the-arts: lightweight U-Net \cite{Sevastopolsky_PRIA_2017}, MNet \cite{Fu_TMI_2018}, JointRCNN \cite{JointRCNN_TBME_2019}, AG-Net \cite{zhang2019AGnet}, FC-DenseNet \cite{DenseNet_Systems_2018},  SAN \cite{Liu_NC_2019}, and HRNet \cite{HRNet}. The first six methods are designed for OD and OC segmentation and the last one is originally designed for natural scene image segmentation. The results of lightweight U-Net \cite{Sevastopolsky_PRIA_2017} and MNet \cite{Fu_TMI_2018} are from \cite{Fu_TMI_2018}. JointRCNN \cite{JointRCNN_TBME_2019}, AG-Net \cite{zhang2019AGnet}, FC-DenseNet \cite{DenseNet_Systems_2018} and SAN \cite{Liu_NC_2019} are from the original paper. Results of HRNet \cite{HRNet} is obtained by fine-tuning with ORGIA \cite{ORIGA}. Following \cite{Fu_TMI_2018} \cite{Liu_NC_2019}, we use overlapping error $E(\hat{Y}, Y) = 1-\frac{Area(\hat{Y} \cap Y)}{Area(\hat{Y} \cup Y)}$ as evaluate metric. Results are reported in Table \ref{tab:PerformanceComparisonORIGA}. As we can see that our method achieves superior segmentation performances to the state-of-the-arts. Fig. \ref{fig:exmaple_OC_ORIGA} and Fig. \ref{fig:exmaple_OD_ORIGA} show two examples for OC and OD segmentation respectively, in which the results by our BoostNet are much closer than MNet \cite{Fu_TMI_2018} and SAN \cite{Liu_NC_2019} to ground-truth.
\begin{table}[!t]
	\begin{center}
		\caption{Performances of different methods on ORIGA \cite{ORIGA}.}\label{tab:PerformanceComparisonORIGA}
		\begin{tabular}{ c | c c c }
\hline {Methods} & ~$E_{disc}$~ & ~$E_{cup}$~ & ~$E_{rim}$~ \\		 \hline  
lightweight U-Net \cite{Sevastopolsky_PRIA_2017}  &  0.115    &  0.287  &   0.303 \\
MNet \cite{Fu_TMI_2018}  &  0.071    &  0.230    &  0.233  \\
JointRCNN \cite{JointRCNN_TBME_2019} & 0.063	 &   	0.209  &  	-\\ 
AG-Net \cite{zhang2019AGnet}  &  0.061    &  0.212   &  - \\
CE-Net \cite{gu2019cenet} & 0.058 & - & -\\
SAN \cite{Liu_NC_2019}  &  0.059    &  0.208   &  0.215 \\ 
HRNet \cite{HRNet}  &  0.057    &  0.209   &  0.208 \\ \hline
\textbf{ours}   & \color {blue}{\textbf{0.051}} &  \color {blue}{\textbf{0.202}} &  \color {blue}{\textbf{0.196}} \\ \hline
		\end{tabular}
	\end{center}
\end{table}

\begin{figure}[!h]
	\centering
	\includegraphics[width=0.9\linewidth]{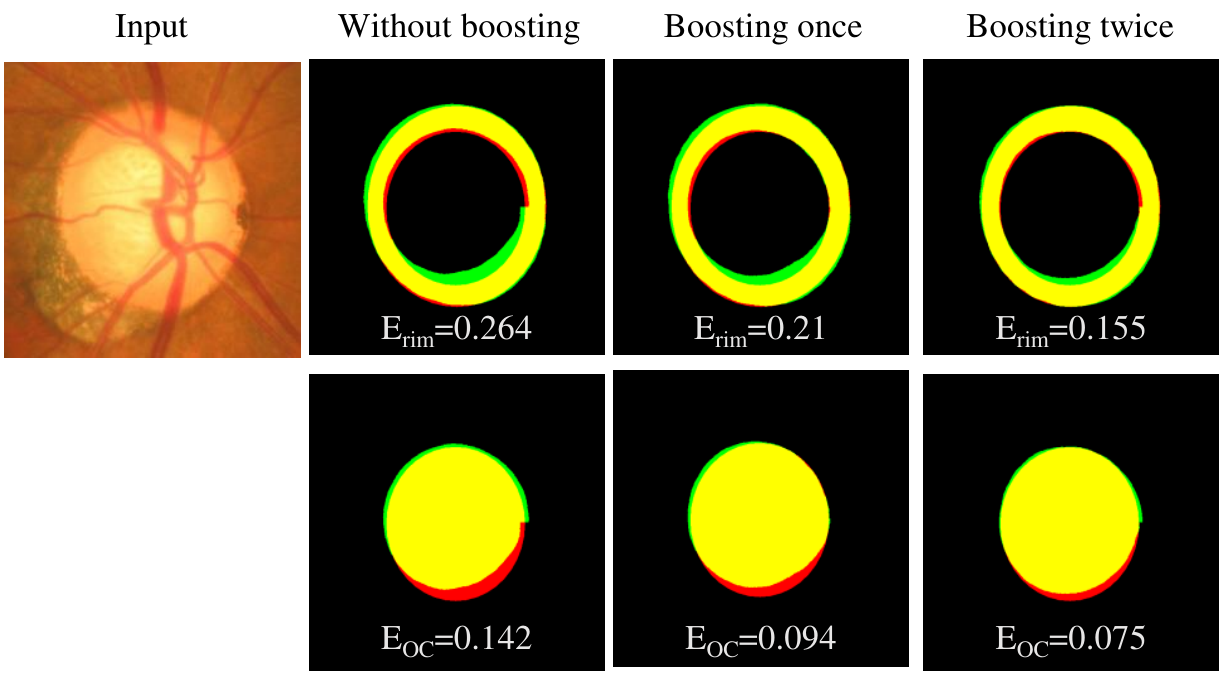}
	\caption{Example illustration for the effectiveness of proposed boosting module. From left to right: input image, results by architectures without boosting, with once boosting and twice boosting. From top to down: results of rim and OC. Pixels in yellow, red and green are correct, miss and error detections respectively.}
	\label{fig:motivation}
\end{figure}

\indent \textbf{Ablation Study.} Table \ref{tab:BMComparisonORIGA} evaluates the effectiveness of boosting. After boosted once, the overlapping errors of OD, OC and rim segmentation are improved from $(0.056, 0.209, $ $0.206)$  to $(0.054, 0.202, 0.200)$. After boosted twice, the performances are improved to $(0.051, 0.202, 0.196)$. Fig. \ref{fig:motivation} shows the results of one example by different segmentation architectures. With boosting, pixels near edges are predicted more accurately, which results in lower overlapping errors.
\begin{table}[!h]
	\begin{center}
		\caption{Ablation study on boosting on ORIGA \cite{ORIGA}.}\label{tab:BMComparisonORIGA}
		\begin{tabular}{ c | c c c }
			\hline {Methods} & ~$E_{disc}$~ & ~$E_{cup}$~ & ~$E_{rim}$~ \\		 \hline  
			without boosting  &  0.056    &  0.209  &   0.206 \\
			boosting once  &  0.054    &  0.202    &  0.200  \\			
			\textbf{boosting twice}   & \color {blue}{\textbf{0.051}} &  \color {blue}{\textbf{0.202}} &  \color {blue}{\textbf{0.196}} \\ \hline
		\end{tabular}
	\end{center}
\end{table}
%

\section{CONCLUSION}
This paper proposes a BoostNet which solves the segmentation problem in a gradient boosting framework. Deformable side-output unit (DSU) and aggregation unit (AU) with deep supervision are designed to estimate the base functions and expansion coefficients in gradient boosting framework. Our BoostNet can be trained end-to-end by standard back-propagation. Validation on ORIGA \cite{ORIGA} demonstrates the effectiveness of our proposed BoostNet on the segmentation of OD and OC in fundus images.
\bibliographystyle{IEEEbib}
\bibliography{strings,refs}

\end{document}